\PassOptionsToPackage{table}{xcolor}

\documentclass[10pt,twocolumn,letterpaper]{article}

\usepackage[pagenumbers]{cvpr}

\usepackage{pifont}
\usepackage{wrapfig}
\newcommand{\cmark}{\ding{51}}

\usepackage{multirow}
\usepackage{xcolor}
\usepackage{amsmath}
\usepackage{graphicx}

\definecolor{colorf}{HTML}{FF9396}
\definecolor{colors}{HTML}{FFC991}
\definecolor{colort}{HTML}{FFF6A9}
\newcommand{\boxcolorf}[1]{%
  \begingroup\setlength{\fboxsep}{1pt}%
  \colorbox{colorf}{\hspace*{2pt}\vphantom{Ay}#1\hspace*{2pt}}%
  \endgroup
}
\newcommand{\boxcolors}[1]{%
  \begingroup\setlength{\fboxsep}{1pt}%
  \colorbox{colors}{\hspace*{2pt}\vphantom{Ay}#1\hspace*{2pt}}%
  \endgroup
}
\newcommand{\boxcolort}[1]{%
  \begingroup\setlength{\fboxsep}{1pt}%
  \colorbox{colort}{\hspace*{2pt}\vphantom{Ay}#1\hspace*{2pt}}%
  \endgroup
}

\definecolor{cvprblue}{rgb}{0.21,0.49,0.74}
\usepackage[pagebackref,breaklinks,colorlinks,citecolor=cvprblue]{hyperref}

\title{Image-Conditioned 3D Gaussian Splat Quantization}

\author{
Xinshuang Liu \quad Runfa Blark Li \quad Keito Suzuki \quad Truong Nguyen \\
University of California, San Diego \\
San Diego, CA, USA \\
{\tt\small xil235@ucsd.edu \quad rul002@ucsd.edu \quad k3suzuki@ucsd.edu \quad tqn001@ucsd.edu} \\
\url{https://XinshuangL.github.io/ICGS-Quantizer}
}

\newif\iffixrefs
\fixrefsfalse

\begin{document}
\maketitle

\begin{abstract}
3D Gaussian Splatting (3DGS) has attracted considerable attention for enabling high‑quality real‑time rendering.
Although 3DGS compression methods have been proposed for deployment on storage‑constrained devices, two limitations hinder archival use: 
(1) they compress medium‑scale scenes only to the megabyte range, which remains impractical for large‑scale scenes or extensive scene collections; and (2) they lack mechanisms to accommodate scene changes after long‑term archival.
To address these limitations, we propose an Image‑Conditioned Gaussian Splat Quantizer (\textbf{ICGS‑Quantizer}) that substantially enhances compression efficiency and provides adaptability to scene changes after archiving. 
ICGS‑Quantizer improves quantization efficiency by jointly exploiting inter‑Gaussian and inter‑attribute correlations and by using shared codebooks across all training scenes, which are then fixed and applied to previously unseen test scenes, eliminating the overhead of per‑scene codebooks.
This approach effectively reduces the storage requirements for 3DGS to the kilobyte range while preserving visual fidelity.
To enable adaptability to post‑archival scene changes, ICGS‑Quantizer conditions scene decoding on images captured at decoding time. 
The encoding, quantization, and decoding processes are trained jointly, ensuring that the codes---quantized representations of the scene---are effective for conditional decoding.
We evaluate ICGS‑Quantizer on 3D scene compression and 3D scene updating. Experimental results show that ICGS‑Quantizer consistently outperforms state‑of‑the‑art methods in compression efficiency and adaptability to scene changes. Our code, model, and data will be publicly available on GitHub.
\end{abstract}

\section{Introduction}

3D Gaussian Splatting (3DGS)~\cite{DBLP:journals/tog/KerblKLD23} has gained considerable attention for its real-time rendering capabilities and high-quality visual outputs. However, 3DGS representations typically require substantial storage, which limits their applicability to large-scale scenes and extensive scene collections. This storage overhead is particularly problematic on storage-constrained devices such as robots, AR/VR headsets, and smartphones.
A promising solution is to compress 3D Gaussians before storage. Recent state-of-the-art methods~\cite{DBLP:conf/eccv/NavaneetMKP24,DBLP:conf/cvpr/LeeRSKP24} achieve compression by quantizing 3D Gaussians into discrete codes and storing associated codebooks for decoding. These codebooks are learned individually per scene via vector quantization~\cite{DBLP:conf/nips/OordVK17,DBLP:conf/cvpr/LeeKKCH22}. While effective, these approaches face two fundamental limitations, which we address with \textbf{ICGS-Quantizer}—an Image-Conditioned Gaussian Splat Quantizer.

\textbf{Limitation 1: Inefficient storage at scale.} Existing 3DGS quantizers typically reduce medium-scale scenes only to the megabyte range, which remains impractical for large-scale environments or extensive collections. Two key factors drive this inefficiency: (1) they train separate codebooks for each scene, incurring considerable storage overhead because each scene must store its own floating-point codebooks in addition to the quantized codes; and (2) they quantize each Gaussian and its attributes (\textit{e.g.}, rotation, scale) independently, missing correlations both across Gaussians and among attributes.
To address this, we train ICGS‑Quantizer on large‑scale data to learn \emph{shared} codebooks across all training scenes, then fix and apply them to previously unseen test scenes, eliminating per‑scene codebook storage. Furthermore, we \emph{jointly} compress all Gaussians and their attributes to capture inter-Gaussian and inter-attribute correlations. Precisely, the scene is partitioned into sparse 3D blocks, each containing multiple Gaussians, and each block is compressed into residual discrete codes. Within each block—which preserves local spatial structure—the Gaussians and their attributes are jointly encoded into latent representations before quantization.
This design reduces 3DGS storage requirements to the kilobyte range while preserving high visual fidelity.

\begin{figure*}
    \centering
    \includegraphics[width=\linewidth]{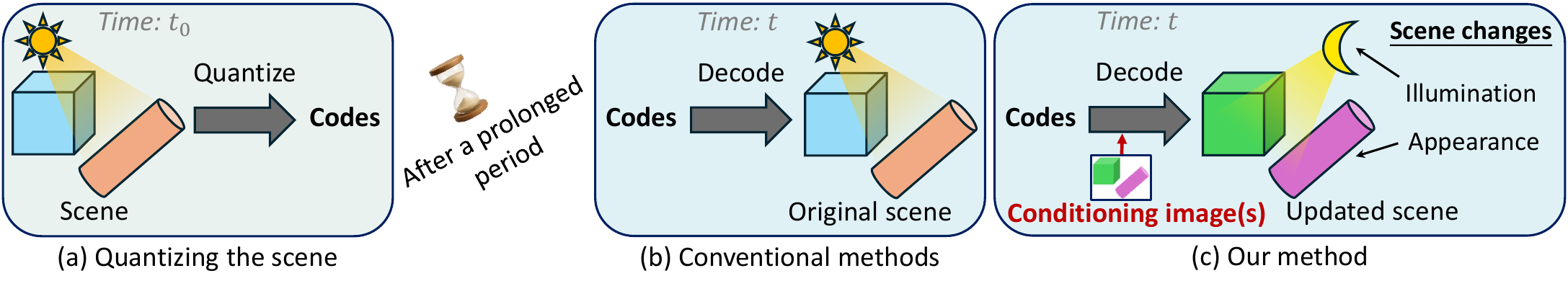}
    \vspace{-15pt}
    \caption{
        \textbf{Image-conditioned scene quantization.}
        (a) At time $t_0$, the scene is encoded and quantized as discrete codes. However, after a prolonged period, the scene may have changed.
        (b) Conventional methods decode the scene from the codes, but can only recover the original scene at time $t_0{<}t$. 
        (c) Our method decodes the scene from the codes conditioned on its image(s) captured at time $t$, adapting the scene to its current illumination and appearance. We recommend watching our videos to observe the dynamic results.
    }
    \vspace{-10pt}
    \label{fig:intro_image_condition}
\end{figure*}

\textbf{Limitation 2: No post-archival scene adaptation.} In real-world applications, achieving a high compression ratio for 3D scenes is not the sole objective. Over time, scenes may change after archival—for example, lighting can shift from day to night, or colors can fade. Therefore, a critical goal during scene decoding is to adapt scenes to their \emph{current} illumination and appearance.
We achieve this by conditioning scene decoding on one or a few current images (Figure~\ref{fig:intro_image_condition}). Specifically, our conditional decoder extracts multi-scale image features from a pre-trained DINOv2 model~\cite{DBLP:journals/tmlr/OquabDMVSKFHMEA24} and fuses them into sparse 3D latents via visibility-aware, coarse-to-fine aggregation.
This design enables the decoded scene to adapt to its current illumination and appearance. Furthermore, the encoding, quantization, and decoding processes are trained jointly, so that the quantized codes are optimized for conditional decoding. When no scene update is needed, our ICGS-Quantizer can still decode the archived scene without requiring any conditioning images.

We evaluate ICGS-Quantizer on both 3D scene compression and 3D scene updating.
For compression, our shared-codebook and joint quantization strategies achieve higher compression ratios and superior rendering quality compared to state-of-the-art methods; for fairness, we do not use conditioning images for our method in this setting.
For scene updating, conditioning on current views provides strong adaptability to changes, achieving the highest rendering quality for updated scenes. 
Additional experiments validate the effectiveness of our coarse-to-fine image-conditioning strategy and demonstrate that the quantizer effectively exploits multiple views while maintaining robustness with only a few input views.

Our contribution can be summarized as follows:
\begin{itemize}    
    \item \textbf{Shared codebooks for all scenes.} We learn codebooks that are shared across all training scenes and fix them at test time, eliminating per-scene codebook storage and improving generalization.
    \item \textbf{Joint Gaussian–attribute quantization.} We jointly quantize multiple Gaussians and their attributes to exploit both inter-Gaussian and inter-attribute correlations, enabling highly efficient compression.    
    \item \textbf{Image-conditioned scene decoding.} We condition scene decoding on a few—or even a single—current view of the scene to adapt it to its current state, in a coarse-to-fine manner. The encoding, quantization, and decoding processes are trained jointly, so that the quantized codes are optimized for conditional decoding.
\end{itemize}

\begin{figure*}
    \centering
    \includegraphics[width=\linewidth]{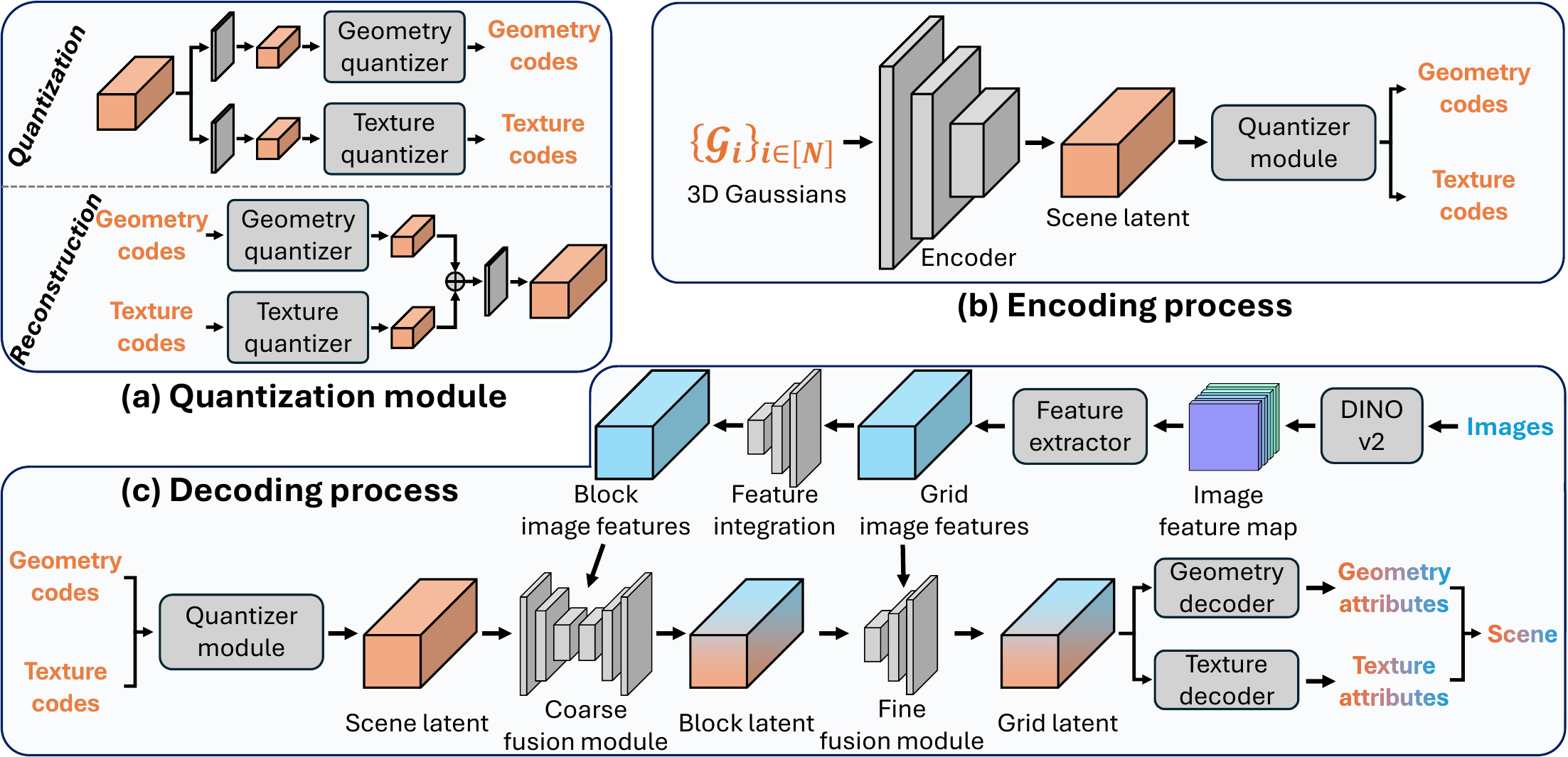}
    \vspace{-15pt}
    \caption{
        \textbf{An overview of our proposed ICGS-Quantizer.}
        (a) Quantization module: geometry and texture are decoupled and quantized independently.
        (b) Joint quantization of 3D Gaussians and their associated attributes. 
        (c) Image-conditioned decoding of 3D Gaussians. Auxiliary neural network branches and training objectives are omitted here for simplicity.
    }
    \vspace{-10pt}
    \label{fig:overview}
\end{figure*}

\section{Related Work}

\subsection{Vector Quantization}

Vector quantization (VQ)~\cite{buzo1980speech,gray1984vector} approximates a vector with its nearest entry in a learned codebook of representative vectors. The objective of VQ is to minimize the discrepancy between the original vectors and their quantized counterparts. 
Various extensions have been proposed to enhance quantization accuracy and efficiency. Residual quantization~\cite{DBLP:conf/icassp/JuangG82,chen2010approximate,martinez2014stacked} iteratively quantizes residuals across multiple stages, with each stage quantizing the difference between the original vector and its reconstruction from previous stages. Product quantization~\cite{sabin2003product,DBLP:journals/pami/JegouDS11,DBLP:journals/pami/GeHK014} partitions vectors into sub-vectors, quantizing each independently.
However, traditional vector quantization methods remain inherently non-differentiable, restricting their compatibility with end-to-end neural network training. 
VQ-VAE~\cite{DBLP:conf/nips/OordVK17} addressed this by copying gradients from the quantizer's outputs to its inputs. 
To enhance accuracy, subsequent work extended this approach as residual quantization~\cite{DBLP:conf/cvpr/LeeKKCH22} and proposed implicit codebooks that are dynamically generated based on the outputs of preceding quantization stages~\cite{DBLP:conf/icml/HuijbenDMSV24}.
\textit{We adopt residual quantization with shared codebooks across all training scenes, fixing them when applied to unseen test scenes. This design achieves high compression by avoiding per-scene codebook storage.}

\subsection{3D Gaussian Splat Compression}

Neural Radiance Fields (NeRFs)~\cite{DBLP:journals/cacm/MildenhallSTBRN22} deliver high-quality renderings and have supported many applications~\cite{DBLP:conf/iccv/WangHHDTL23,DBLP:conf/cvpr/MetzerRPGC23,liu2025image}. Subsequent methods enhanced NeRF's computational efficiency by using ensembles of smaller MLPs~\cite{DBLP:conf/iccv/ReiserPL021,DBLP:journals/tog/MullerESK22}, each operating efficiently. Other methods enhanced sampling efficiency through grid-based~\cite{DBLP:conf/cvpr/Fridovich-KeilY22} or point-based~\cite{DBLP:conf/cvpr/XuXPBSSN22} representations. 
Combining both computational and sampling efficiency, 3D Gaussian Splatting (3DGS)~\cite{DBLP:journals/tog/KerblKLD23} enables fast 3D reconstruction and real-time rendering, forming a computationally efficient foundation for various tasks~\cite{DBLP:conf/cvpr/GuedonL24,li2024splatsdf,DBLP:conf/cvpr/QianWM0024}. 
However, unlike NeRF-based methods, 3DGS incurs substantial storage due to the large number of Gaussians. To mitigate this, recent methods~\cite{DBLP:conf/eccv/ChenWLHC24,DBLP:conf/cvpr/0005YXX0L024,DBLP:conf/nips/WangLGYKW24} proposed compact 3DGS representations by leveraging anchor levels that capture shared features among Gaussians. Further compression methods~\cite{DBLP:conf/nips/FanWWZXW24,DBLP:conf/cvpr/LeeRSKP24,DBLP:conf/cvpr/NiedermayrSW24,DBLP:conf/eccv/NavaneetMKP24,DBLP:conf/eccv/GirishGS24} applied vector quantization (VQ)~\cite{DBLP:conf/nips/OordVK17,DBLP:conf/cvpr/LeeKKCH22} to quantize Gaussians into discrete codes alongside necessary floating-point data, reducing storage demands to the megabyte range. However, these approaches typically (i) train unshared, per-scene codebooks, adding floating-point storage overhead, and (ii) quantize each Gaussian and its attributes independently, overlooking inter-Gaussian and inter-attribute correlations. 
\textit{In contrast, we learn shared codebooks across training scenes and jointly quantize multiple Gaussians with their attributes, enabling kilobyte-level compression while preserving visual fidelity, thus enhancing scalability for large or numerous scenes on storage-limited devices.}

\subsection{3D Scene Updating}

Several scene editing methods have been proposed for 3D Gaussian Splatting (3DGS)~\cite{DBLP:journals/tog/KerblKLD23}, enabling applications such as object removal~\cite{DBLP:conf/eccv/WangWZX24}, scene relighting~\cite{DBLP:conf/mm/GuoBHGLC0024,DBLP:conf/eccv/GaoGLLZCZY24,DBLP:conf/siggrapha/Bi00PF0W24}, and natural language-driven edits~\cite{DBLP:conf/eccv/ChenLV24,DBLP:conf/cvpr/ChenCZWYWCYLL24,DBLP:conf/eccv/WangYWZCZ24,DBLP:conf/eccv/WuBLWRTP24}.
While these methods produce high-quality results, they do not address our objective: adapting scenes to their current views when decoding them from previously quantized codes.
The most relevant work lies in style transfer~\cite{DBLP:conf/cvpr/DengTDMPWX22,DBLP:conf/iccv/HongJLAKLKUB23,DBLP:conf/cvpr/ChungHH24}, which adjusts the style of one image to match reference images.
Deng \etal~\cite{DBLP:conf/cvpr/DengTDMPWX22} utilized transformer architectures for high-quality style transfer. Hong \etal~\cite{DBLP:conf/iccv/HongJLAKLKUB23} introduced a pattern repeatability metric to quantify the rhythm and repetition of local patterns in the style image, thereby enhancing style transfer efficacy. Chung \etal~\cite{DBLP:conf/cvpr/ChungHH24} leveraged pre-trained diffusion models to enhance style transfer performance. Recent works~\cite{DBLP:conf/siggrapha/LiuZXT0L24,jain2024stylesplat} incorporated style transfer into 3DGS for multi-view consistency.
\textit{Unlike these methods, which focus on style manipulation, we perform image-conditioned decoding of previously quantized scene codes, enabling scene updates to their current views without re-optimization.}

\section{Methodology}
\label{sec:method}

Figure~\ref{fig:overview} provides an overview of our method.
We first encode each 3DGS scene into a latent representation that is decoupled into geometry and texture features (Sec.~\ref{subsec:encode}), with this decoupling reinforced by the training objectives in Sec.~\ref{subsec:loss}.
The resulting features are quantized into discrete geometry and texture codes via residual vector quantizers (RVQs) whose codebooks are \emph{shared across all training scenes} and frozen at test time, eliminating per‑scene codebook storage.
Storing only discrete codes rather than floating‑point latent vectors significantly reduces storage.
To reconstruct a scene (Sec.~\ref{subsec:decode}), we recover the latent representation from the codes and decode Gaussian attributes, optionally conditioning on current images to adapt the scene to its current illumination and appearance.

\subsection{Preliminary}

\textbf{3D Gaussian Splatting.} 
Kerbl \etal~\cite{DBLP:journals/tog/KerblKLD23} proposed representing a scene using a collection of 3D Gaussians. Each 3D Gaussian $\mathcal{G}$ is parameterized as follows: (1) a mean vector $\boldsymbol{\mu} \in \mathbb{R}^3$ indicating its center; (2) spherical harmonics (SH) coefficients $\boldsymbol{y} \in \mathbb{R}^k$ modeling view-dependent appearance, with $k$ denoting the degrees of freedom; (3) a quaternion $\boldsymbol{r} \in \mathbb{R}^4$ for rotation; (4) a scaling factor $\boldsymbol{s} \in \mathbb{R}^3$; and (5) an opacity value $\sigma \in \mathbb{R}_+$. A scene is thus defined by a collection $\{ \mathcal{G}_i = ( \boldsymbol{\mu}_i, \boldsymbol{y}_i, \boldsymbol{r}_i, \boldsymbol{s}_i, \sigma_i ) \}_{i \in [N]}$, where $N$ denotes the number of Gaussians. 
Using the rotation matrix $\boldsymbol{R}$ derived from $\boldsymbol{r}$, the Gaussian covariance $\boldsymbol{\Sigma} \in \mathbb{R}^{3 \times 3}$ is computed as:
\begin{equation}
\boldsymbol{\Sigma} = \boldsymbol{R}\operatorname{diag}(\boldsymbol{s})^2\boldsymbol{R}^T.
\end{equation}
Given a viewing transformation $\boldsymbol{W}$ and the Jacobian $\boldsymbol{J}$ of the affine approximation of the projective transformation, Zwicker \etal~\cite{DBLP:conf/siggraph/ZwickerPBG01} compute the covariance in camera coordinates as:
\begin{equation}
\boldsymbol{\Sigma}' = \boldsymbol{J}\boldsymbol{W}\boldsymbol{\Sigma}\boldsymbol{W}^T\boldsymbol{J}^T.
\end{equation}
The color of each rendered pixel $\boldsymbol{p}$ is computed as:
\begin{equation}
\boldsymbol{C}(\boldsymbol{p}) = \sum^N_{i=1} \boldsymbol{c}_i \sigma_i G_i^{2D}(\boldsymbol{p}) \prod_{j=1}^{i-1} \left( 1 - \sigma_j G_j^{2D}(\boldsymbol{p}) \right),
\end{equation}
where $\boldsymbol{c}_i$ denotes the view-dependent color computed from the SH coefficients $\boldsymbol{y}_i$, and $G_i^{2D}(\boldsymbol{p})$ is defined as:
\begin{equation}
G_i^{2D}(\boldsymbol{p}) = e^{-\frac{1}{2} (\boldsymbol{p} - \boldsymbol{\mu}_i)^T \left( \boldsymbol{\Sigma}^{2D} \right)^{-1} (\boldsymbol{p} - \boldsymbol{\mu}_i)},
\end{equation}
where $\boldsymbol{\Sigma}^{2D}$ is derived by removing the third row and column of $\boldsymbol{\Sigma}'$.

\textbf{Vector Quantization.} 
Vector quantization (VQ) uses a codebook $\mathcal{C} = \{(i,\boldsymbol{e}_i)\}_{i \in [N]}$, containing $N$ codes $i$ and associated embeddings $\boldsymbol{e}_i \in \mathbb{R} ^ {d_e}$. For an input vector $\boldsymbol{x} \in \mathbb{R}^{d_e}$, VQ outputs the nearest code embedding:
\begin{equation}
\mathcal{Q}(\boldsymbol{x};\mathcal{C}) = \mathop{\arg\min}_{\substack{i \in [N]}} \left\| \boldsymbol{x} - \boldsymbol{e}_i \right\|_2^2.
\end{equation}
To reduce the quantization error, residual vector quantization (RVQ) quantizes the input vector using $D$ codebooks. Starting with $\boldsymbol{r}_0 = \boldsymbol{x}$, RVQ quantizes the residual value $\boldsymbol{r}_d$ at each iteration $d$:
\begin{equation}
\begin{aligned}
    i_d &= \mathcal{Q}(\boldsymbol{r}_{d-1}; \mathcal{C}_d), \\
    \boldsymbol{r}_d &= \boldsymbol{r}_{d-1} - \boldsymbol{e}_{i_d},
\end{aligned}
\end{equation}
for $d=1, \dots, D$. The final approximation $\boldsymbol{\hat{x}}$ is obtained by summing the selected code embeddings:
\begin{equation}
\boldsymbol{\hat{x}} = \sum_{d=1}^{D} \boldsymbol{e}_{i_d}.
\end{equation}
RVQ requires $D \log_2 N$ bits of storage per input and offers exponentially increasing representational capacity with respect to the number of codebooks, serving as the foundation of our method.

\subsection{Joint Quantization of Gaussians and Attributes}
\label{subsec:encode}

We encode a 3DGS scene into latent features and quantize them into discrete codes for efficient storage.

\textbf{Grid 3DGS representation.}
Optimizing 3DGS typically requires Gaussian pruning/splitting to avoid poor local minima, making the number of Gaussians difficult to control and complicating storage planning. To address this, we discretize space into a sparse voxel grid; each non‑empty cell holds one Gaussian and empty cells are omitted. This removes the need to store floating‑point Gaussian centers $\boldsymbol{\mu}$ and enables direct control of storage via grid resolution.

\textbf{Grid‑block feature encoding.}
As illustrated in Figure~\ref{fig:overview}(b), we encode the per‑grid Gaussian attributes—SH coefficients $\boldsymbol{y}$, quaternion $\boldsymbol{r}$, scale $\boldsymbol{s}$, and opacity $\sigma$—into vectors $\boldsymbol{f}_y, \boldsymbol{f}_r, \boldsymbol{f}_s, \boldsymbol{f}_\sigma \in \mathbb{R}^{d_g}$, where $d_g{=}32$ is the grid-level feature dimension. These vectors are concatenated and linearly projected to form a grid feature vector $\boldsymbol{f}_g \in \mathbb{R}^{d_g}$.
To capture local structure, we group each $K \times K \times K$ neighboring grids into a block (with $K{=}4$). Within each block, grid feature vectors are processed by 3D convolutional layers to produce a block feature vector $\boldsymbol{f}_b \in \mathbb{R}^{d_b}$ ($d_b{=}128$), padding empty grids with a learnable vector. 
The feature vectors from all non‑empty blocks form a sparse 3D feature map, which a 3D U‑Net~\cite{DBLP:conf/miccai/RonnebergerFB15} further processes—again with learned padding—to produce the scene’s sparse latent.

\textbf{Scene latent quantization.}
We project the latent into geometry and texture features and quantize each with a RVQ to obtain geometry and texture codes (Figure~\ref{fig:overview}(a)). The training objectives in Sec.~\ref{subsec:loss} enforce the intended decoupling. 
To eliminate per-scene codebook storage and enhance generalizability, we use shared codebooks across all training scenes, and fix them for unseen scenes at test time.

Our method reduces 3DGS storage to the kilobyte range while preserving high visual fidelity by (i) jointly encoding Gaussians to exploit correlations and (ii) sharing codebooks across scenes to avoid per-scene codebook storage.

\subsection{Image-Conditioned Scene Decoding}
\label{subsec:decode}

We decode a scene from its codes and optionally condition the process on its current images to adapt to scene changes.

\textbf{Scene latent recovery.}
The decoding process begins by reconstructing geometry and texture latents from their codes via the RVQs and fusing them with a linear projection to form a scene latent (Figure~\ref{fig:overview}(a)). 

\textbf{Image feature extraction.}
Optionally, in order to adapt to scene changes, we condition the scene decoding on features extracted from one or more current images (Figure~\ref{fig:overview}(c)). Image features come from a frozen DINOv2 model~\cite{DBLP:journals/tmlr/OquabDMVSKFHMEA24} augmented by a lightweight 2D U-Net (trained from scratch). Grid-level features are obtained by projecting non-empty grid centers onto image planes to query the feature maps, then averaging the retrieved features for each grid cell, weighted by view-dependent visibility; to obtain robust weights, we estimate visibility using non-empty cells only and share the same visibility across cells in a block. The resulting grid‑level features are then aggregated to form block‑level image features.

\textbf{Coarse-to-fine conditioning.}
When images are available, scene decoding is conditioned at two levels. \emph{Coarse image condition:} block‑level image features are injected into the scene latent via a 3D U‑Net to produce a block latent, with empty blocks padded by a learnable vector. \emph{Fine image condition:} the block latent is upsampled to the grid cells and fused with grid-level image features to produce a grid latent. This coarse‑to‑fine design first imparts global semantics and then refines local details. The effectiveness of both conditioning stages is validated in Sec.~\ref{subsec:exp_ablation}.

\textbf{Gaussian attribute decoding.}
Finally, the Gaussian attributes are decoded from the grid latent using geometry and texture decoders. Geometry attributes include the rotation quaternion $\boldsymbol{r}$, scaling factor $\boldsymbol{s}$, and opacity $\sigma$; texture attributes include spherical harmonics coefficients $\boldsymbol{y}$.

In summary, scene decoding can be optionally conditioned on image features to adapt to current illumination and appearance. This conditioning follows a coarse-to-fine strategy: the coarse stage imparts global semantics, while the fine stage refines local details.

\subsection{Training Objective}
\label{subsec:loss}

We supervise the model using both Gaussian attributes, which provide direct guidance early in training, and images, which enhance the final visual quality.

\textbf{Loss functions.}
For Gaussian attribute supervision, we use a Huber loss~\cite{huber1992robust} on rotations and scaling factors and a binary cross‑entropy on opacity. All terms use equal weights of $1$ without dataset‑specific tuning, minimizing hyperparameters and enhancing reproducibility. The Huber loss stabilizes training by minimizing sensitivity to outliers.
For image supervision, we adopt the photometric loss from 3DGS~\cite{DBLP:journals/tog/KerblKLD23} and add a Mean Squared Error (MSE) term for training stability.
To ensure compatibility of vector quantization within the deep neural network, we additionally include a commitment loss term~\cite{DBLP:conf/nips/OordVK17}. 

\textbf{Training objective design}.
We supervise the model both \emph{before} image conditioning (to preserve the archived scene) and \emph{after} conditioning (to adapt to the current scene), thereby optimizing the quantized codes for both archival fidelity and post-archival adaptability. Geometry–texture decoupling is enforced through (i) separate quantizers and decoders for geometry and texture features, and (ii) a stop-gradient from non-geometry objectives into the geometry quantizer outputs, encouraging geometry codes to specialize in geometric signals. 

\textbf{Optimization.}
We train the model for 100 epochs using AdamW~\cite{DBLP:conf/iclr/LoshchilovH19} ($\beta_1{=}0.9$, $\beta_2{=}0.95$) with gradient clipping and a batch size of $8$.
In epochs 1–10, vector quantization is disabled and the learning rate is linearly warmed up~\cite{DBLP:conf/nips/VaswaniSPUJGKP17} to $1{\times}10^{-4}$.  
In epochs 11–90, the learning rate follows a cosine decay schedule~\cite{DBLP:conf/iclr/LoshchilovH17} down to $1{\times}10^{-5}$. In the same interval, the Gaussian-attribute loss weight is reduced from $0.1$ to $0$, while the commitment loss weight is increased from $0.01$ to $0.1$. Codebooks are updated via $k$-means clustering every two epochs.  
In the final 10 epochs, both the learning rate and codebooks are fixed.

This subsection presents the training objective; the complete formulation of the losses and additional implementation details are provided in the supplementary material.

\begin{table*}
  \caption{\textbf{Comparison with state-of-the-art methods for 3D scene compression.} \#Images denotes the number of images used during testing.
  The results are highlighted as \boxcolorf{best}, \boxcolors{second-best}, and \boxcolort{third-best}, respectively.}
  \label{tab:compression}
  \centering
  \begin{tabular}{l|c|cccc}
    \toprule
    Method & \#Images & PSNR $\uparrow$ & SSIM $\uparrow$ & LPIPS $\downarrow$ & Storage (KB) \\
    \midrule
      3DGS-small & 0 & 25.23 & 0.9345 & 0.1546 & 121.0 \\
      3DGS-medium & 0 & \cellcolor{colort}28.79 & \cellcolor{colort}0.9641 & 0.0729 & 561.9 \\
      3DGS-large & 0 & \cellcolor{colorf}30.92 & \cellcolor{colorf}0.9777 & \cellcolor{colorf}0.0292 & 2661.1 \\
    \midrule
    CompGS & 0 & 27.47 & 0.9545 & 0.0830 & \cellcolor{colort}21.9 \\
    C3DGS & 0 & 27.90 & 0.9582 & 0.0785 & \cellcolor{colors}21.1 \\
    \midrule
    \multirow{2}{*}{ICGS-Quantizer} & 0 & 28.50 & 0.9634 & \cellcolor{colort}0.0405 & \cellcolor{colorf}16.9 \\
     & 1 & \cellcolor{colors}29.52 & \cellcolor{colors}0.9681 & \cellcolor{colors}0.0339 & \cellcolor{colorf}16.9 \\
    \bottomrule
  \end{tabular}
\end{table*}

\begin{figure*}
  \centering
  \includegraphics[width=\linewidth]{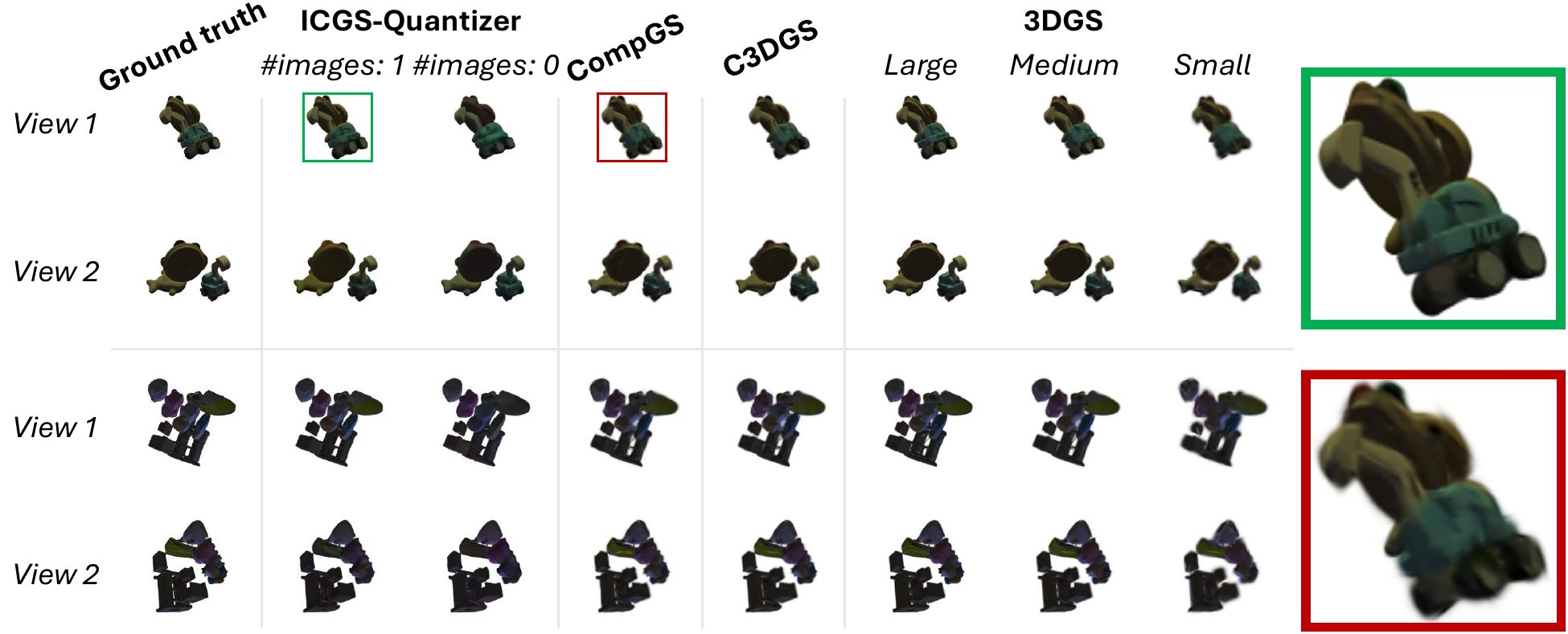}
  \caption{
      \textbf{Qualitative results of 3D scene compression.} Zoom in to examine details. ``Large'',  ``Medium'', and ``Small'' indicate the number of Gaussians, balancing quality and storage.
  }
  \label{fig:compression}
\end{figure*}

\section{Experiments}
\label{sec:experiment}

This section presents a thorough evaluation of ICGS-Quantizer.
Sec.~\ref{subsec:exp_setup} details the experiment setup;
Sec.~\ref{subsec:exp_compression} and Sec.~\ref{subsec:exp_updating} report results for 3D scene compression and 3D scene updating, respectively;
Sec.~\ref{subsec:exp_ablation} analyzes the effect of image conditioning and the number of input images.

\subsection{Experiment Setup}
\label{subsec:exp_setup}

We evaluate our method on the \textit{Google Scanned Objects} (GSO) dataset~\cite{DBLP:conf/icra/DownsFKKHRMV22}, which contains real-world scanned objects. After quality filtering, a total of 360 objects are used to create 36,000 training scenes, while the remaining 100 unseen objects form 100 test scenes for evaluation. Each scene has two states: an archived version at time $t_0$ and a changed version at time $t>t_0$. Details of the dataset construction are provided in the supplementary material.
For quantitative evaluation, we render 50 novel views per scene by uniformly sampling camera poses over a full $360^{\circ}$ rotation.
We report Peak Signal-to-Noise Ratio (PSNR), Structural Similarity Index Measure (SSIM)~\cite{DBLP:journals/tip/WangBSS04}, and Learned Perceptual Image Patch Similarity (LPIPS)~\cite{DBLP:conf/cvpr/ZhangIESW18}. 
All methods operate with 3 degrees of freedom for each Gaussian spherical harmonic. Our method uses residual codebooks with a depth of 4 and a size of 1024, shared across all training scenes. The codebooks remain fixed during evaluation on test scenes.
To contextualize performance under varying storage constraints, we include grid-based 3DGS variants as baselines for both 3D scene compression and 3D scene updating, benefiting from their convenient storage management: \textit{3DGS-small}, \textit{3DGS-medium}, and \textit{3DGS-large}, corresponding to grid resolutions of $32$, $64$, and $128$, respectively.

\subsection{3D Scene Compression}

\label{subsec:exp_compression}

In addition to the 3DGS baselines, we compare against state-of-the-art compression methods for 3DGS representations, including CompGS~\cite{DBLP:conf/eccv/NavaneetMKP24} and C3DGS~\cite{DBLP:conf/cvpr/NiedermayrSW24}, using a codebook size of $256$ per scene. For a fair comparison, both our method and the baseline compression methods are applied to well-optimized 3DGS without image-based re-optimization, with Gaussians having negative opacity logits discarded to reduce storage.
Our method employs a block resolution of $32$ for storage, whereas the compared compression methods use a grid resolution of $64$ for storage, which is $2\times$ finer than ours and thus requires more storage.

Experimental results are presented in Table~\ref{tab:compression} and Figure~\ref{fig:compression}.
As expected, the performance of 3DGS improves with increased storage capacity, underscoring the importance of evaluating compression methods under comparable storage budgets. Under such constraints, our method significantly outperforms the state-of-the-art scene compression approaches even without the aid of conditioning images. For example, C3DGS achieves 27.90 PSNR, 0.9582 SSIM, and 0.0785 LPIPS with 21.1 KB of storage, whereas ICGS-Quantizer, at 16.9 KB, attains 28.50 PSNR, 0.9634 SSIM, and 0.0405 LPIPS. Remarkably, in this setting (no conditioning images), ICGS-Quantizer reaches performance comparable to uncompressed \textit{3DGS-medium}, which requires $33\times$ the storage. With a single conditioning image, ICGS-Quantizer further approaches uncompressed \textit{3DGS-large}, which demands $157\times$ the storage.

These results demonstrate the storage efficiency of our method, attributable to (i) exploiting inter-Gaussian and inter-attribute correlations to reduce redundancy and (ii) using shared codebooks across all scenes, which eliminates per-scene codebook storage.

\begin{table*}
  \caption{\textbf{Comparison with state-of-the-art methods for 3D scene updating.} The results are highlighted as \boxcolorf{best}, \boxcolors{second-best}, and \boxcolort{third-best}, respectively.}
  \label{tab:updating}
  \centering
  \begin{tabular}{cl|ccc}
    \toprule
    Storage (KB) & Method & PSNR $\uparrow$ & SSIM $\uparrow$ & LPIPS $\downarrow$ \\
    \midrule
     \multirow{6.5}{*}{121.0} & Archived 3DGS & 24.01 & 0.9210 & 0.1655 \\
     & Fine-tuned 3DGS & 23.13 & 0.9238 & 0.1325 \\
     & Current 3DGS & 25.23 & 0.9345 & 0.1546 \\
    \cmidrule(r){2-5}

     & StyleID & 23.75 & 0.9177 & 0.1188 \\
     & AesPA-Net & 23.91 & 0.9266 & 0.1141 \\
     & StyTR2 & 23.57 & 0.9252 & 0.1177 \\

    \midrule
     \multirow{6.5}{*}{561.9} & Archived 3DGS & 26.13 & 0.9421 & 0.0895 \\
     & Fine-tuned 3DGS & 26.52 & 0.9495 & 0.0598 \\
     & Current 3DGS & \cellcolor{colort}28.79 & \cellcolor{colort}0.9641 & 0.0729 \\
    \cmidrule(r){2-5}

     & StyleID & 25.46 & 0.9320 & 0.0603 \\
     & AesPA-Net & 26.40 & 0.9435 & 0.0647 \\
     & StyTR2 & 26.27 & 0.9458 & 0.0635 \\

    \midrule
     \multirow{6.5}{*}{2661.1} & Archived 3DGS & 26.92 & 0.9490 & 0.0502 \\
     & Fine-tuned 3DGS & 28.46 & 0.9599 & \cellcolor{colort}0.0312 \\
     & Current 3DGS & \cellcolor{colorf}30.92 & \cellcolor{colorf}0.9777 & \cellcolor{colors}0.0292 \\
    \cmidrule(r){2-5}

     & StyleID & 25.78 & 0.9346 & 0.0383 \\
     & AesPA-Net & 27.49 & 0.9494 & 0.0455 \\
     & StyTR2 & 27.93 & 0.9552 & 0.0414 \\

    \midrule
     16.9 & ICGS-Quantizer & \cellcolor{colors}30.45 & \cellcolor{colors}0.9772 & \cellcolor{colorf}0.0232 \\
    \bottomrule
  \end{tabular}
\end{table*}

\subsection{3D Scene Updating}

\label{subsec:exp_updating}

We evaluate the capability of ICGS-Quantizer to adapt the scene to post-archival changes through a 3D scene updating task. Each scene has two states: one archived at time $t_0$ and one current at time $t$. The archived scene serves as input, while the current scene serves as ground truth. For all methods, six images are used to update the scene.
Since the task requires adapting illumination and appearance from the provided images, we compare ICGS-Quantizer with style transfer methods, including StyleID~\cite{DBLP:conf/cvpr/ChungHH24}, AesPA-Net~\cite{DBLP:conf/iccv/HongJLAKLKUB23}, and StyTR2~\cite{DBLP:conf/cvpr/DengTDMPWX22}.
We average their outputs from each reference to use multiple reference images, and refine the results using masks rendered by 3DGS.
As 3D scene updating relates to continual learning, we further include comparisons with 3DGS under three settings: (1) using the archived 3DGS (a lower-end performance reference), (2) fine-tuning the archived 3DGS with current images (a continual-learning baseline), and (3) using the current 3DGS (an upper-end performance reference).

\begin{figure*}
  \centering
  \includegraphics[width=\linewidth]{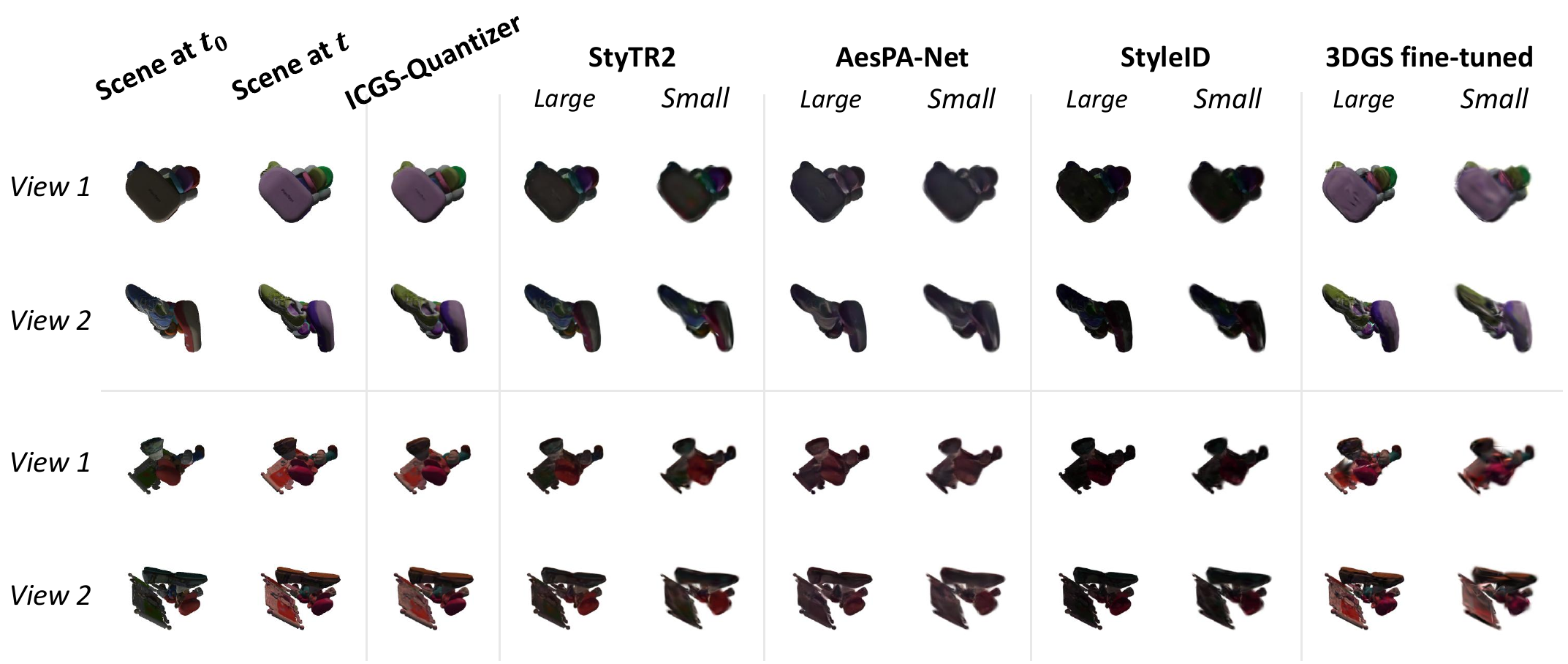}
  \caption{
      \textbf{Qualitative results of 3D scene updating.} For the baseline methods, ``Large'' and ``Small'' refer to configurations utilizing varying numbers of Gaussians, balancing quality and storage.
  }
  \label{fig:updating}
\end{figure*}

Experimental results are presented in Table~\ref{tab:updating} and Figure~\ref{fig:updating}. Our method is compared against baselines with varying storage budgets, all of which require more storage than ours.
Notably, the 3DGS of the archived scene (\textit{Archived 3DGS}) performs poorly, despite utilizing substantial storage resources (2661.1 KB). This is primarily due to the appearance gap between the archived and current scenes, which cannot be bridged by increasing storage without incorporating information from the current scene. In contrast, the 3DGS of the current scene (\textit{Current 3DGS}) performs well when provided with sufficient storage, highlighting the importance of incorporating updated visual information. Fine-tuning the archived 3DGS with current images (\textit{Fine-tuned 3DGS}) improves performance over the \textit{Archived 3DGS}, but still falls short of \textit{Current 3DGS}. This indicates that while reference images contribute positively, the continual learning paradigm is constrained by a limited number of inputs.
Among style transfer methods, StyTR2 at 2661.1 KB achieves the highest PSNR (27.93), comparable to \textit{Fine-tuned 3DGS} (28.46) while requiring no re-optimization. This demonstrates the effectiveness of style transfer in low-data scenarios.
Finally, our ICGS-Quantizer significantly outperforms all baselines and is comparable to the upper-end performance reference (\textit{Current 3DGS} at 2661.1 KB), while surpassing it on LPIPS, yet requiring the least storage and no re-optimization. 

Trained on large-scale scenes, ICGS-Quantizer learns to effectively extract and store the core information needed to decode scenes conditioned on current images. As a result, it demonstrates superior adaptability with minimal storage overhead, successfully updating scenes to reflect post-archival changes.

\begin{table*}
  \caption{\textbf{Ablation study of image conditioning on 3D scene compression and 3D scene updating.} Six images are used for conditioning.}
  \label{tab:ablation}
  \centering
  \begin{tabular}{cc|ccc|ccc}
    \toprule
    & & \multicolumn{3}{c|}{3D scene compression} & \multicolumn{3}{c}{3D scene updating} \\
    Coarse image condition & Fine image condition & PSNR $\uparrow$ & SSIM $\uparrow$ & LPIPS $\downarrow$  & PSNR $\uparrow$ & SSIM $\uparrow$ & LPIPS $\downarrow$ \\
    \midrule
    & & 28.50 & 0.9634 & 0.0405 & 26.33 & 0.9466 & 0.0524 \\
    \cmark & & 30.34 & 0.9752 & 0.0261 & 30.34 & 0.9750 & 0.0262 \\
    \cmark & \cmark & 30.44 & 0.9773 & 0.0232 & 30.45 & 0.9772 & 0.0232 \\
    \bottomrule
  \end{tabular}
\end{table*}

\begin{figure}
    \centering
    \includegraphics[width=\linewidth]{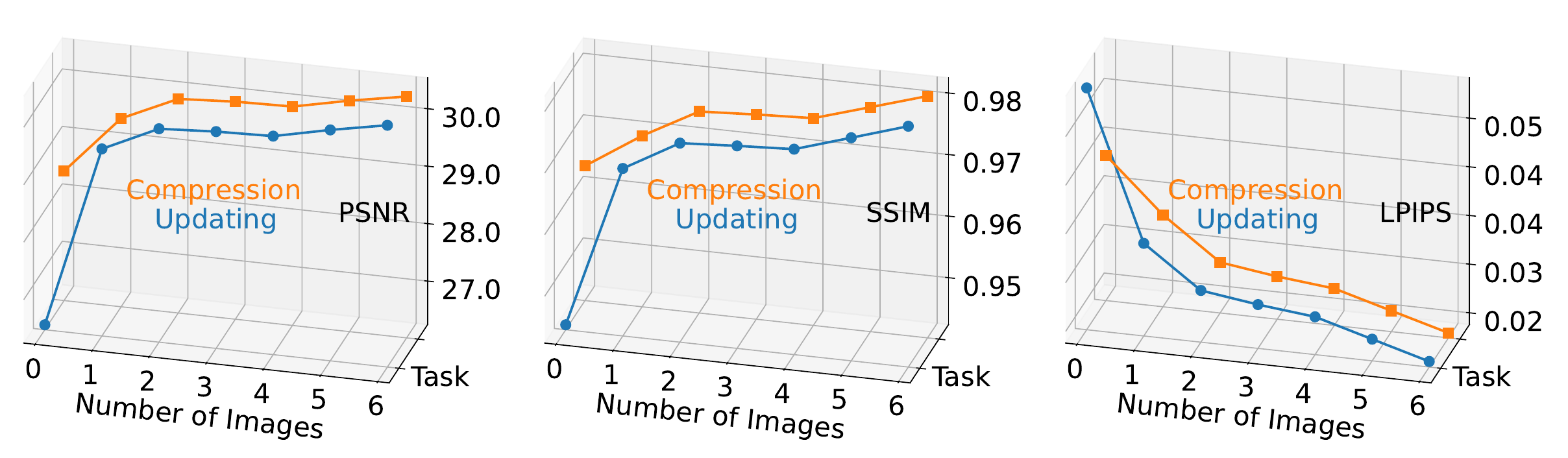}
    \caption{
        Experimental results of ICGS-Quantizer for 3D scene compression (``compression'') and 3D scene updating (``updating'') using varying numbers of input images. Since the conditioning images are optional, ``0'' indicates no image is used.
    }
    \label{fig:result_img_num}
\end{figure}

\subsection{Evaluation of Image Conditioning}

\label{subsec:exp_ablation}

We begin with an ablation study assessing the effectiveness of image conditioning. As shown in Table~\ref{tab:ablation}, incorporating a \textit{coarse image condition} yields a substantial improvement, highlighting the benefit of using up-to-date images during decoding. Further gains are achieved when using a \textit{fine image condition}, with consistent improvements across PSNR, SSIM, and LPIPS for both tasks, thereby validating the effectiveness of integrating image information at a finer level.

We also examine the effect of varying the number of conditioning images. As shown in Figure~\ref{fig:result_img_num}, performance improves steadily as the number of input images increases, as reflected by PSNR, SSIM, and LPIPS. Remarkably, even with a single conditioning image, the model maintains strong performance. These observations demonstrate both the model’s capacity to exploit multi-view information and its robustness in scenarios with limited input views.

\section{Conclusion}

We introduce ICGS‑Quantizer, an Image-Conditioned Gaussian Splat Quantizer for archiving large‑scale scenes and extensive scene collections. ICGS‑Quantizer enhances quantization efficiency by (i) exploiting inter‑Gaussian and inter‑attribute correlations to remove redundancy and (ii) learning shared codebooks across training scenes that are fixed at test time, eliminating per‑scene codebook storage. This design drives storage down to the kilobyte range while preserving visual fidelity. To enable adaptability to post-archival scene changes, ICGS-Quantizer conditions scene decoding on images captured at decoding time. The encoding, quantization, and decoding processes are trained jointly, so that the quantized codes are optimized for conditional decoding. Experimental results demonstrate that ICGS-Quantizer outperforms state-of-the-art methods in both compression efficiency and adaptability to scene changes.

\textbf{Future Work.} This work presents a promising approach for archiving data as discrete codes that are particularly well-suited for conditional decoding. Future research may explore extending this approach to other data modalities, such as images and audio. Additionally, incorporating pre-trained foundation models into the method offers a potential direction for further enhancing compression efficiency.

{
    \small
    \bibliographystyle{ieeenat_fullname}
    \iffixrefs
        \IfFileExists{main.bbl}{%

        }{%
            \typeout{*** no main.bbl：please use \fixrefsfalse，first then use true ***}%
        }
    \else
        \bibliography{main}

\begin{thebibliography}{55}
\providecommand{\natexlab}[1]{#1}
\providecommand{\url}[1]{\texttt{#1}}
\expandafter\ifx\csname urlstyle\endcsname\relax
  \providecommand{\doi}[1]{doi: #1}\else
  \providecommand{\doi}{doi: \begingroup \urlstyle{rm}\Url}\fi

\bibitem[Bi et~al.(2024)Bi, Zeng, Zeng, Pei, Feng, Zhou, and Wu]{DBLP:conf/siggrapha/Bi00PF0W24}
Zoubin Bi, Yixin Zeng, Chong Zeng, Fan Pei, Xiang Feng, Kun Zhou, and Hongzhi Wu.
\newblock {GS}\({}^{\mbox{3}}\): Efficient relighting with triple gaussian splatting.
\newblock In \emph{{SIGGRAPH} Asia 2024 Conference Papers, {SA} 2024, Tokyo, Japan, December 3-6, 2024}, pages 12:1--12:12. {ACM}, 2024.

\bibitem[Buzo et~al.(1980)Buzo, Gray, Gray, and Markel]{buzo1980speech}
Andr{\'e}s Buzo, A Gray, R Gray, and John Markel.
\newblock Speech coding based upon vector quantization.
\newblock \emph{IEEE Transactions on Acoustics, Speech, and Signal Processing}, 28\penalty0 (5):\penalty0 562--574, 1980.

\bibitem[Chen et~al.(2024{\natexlab{a}})Chen, Laina, and Vedaldi]{DBLP:conf/eccv/ChenLV24}
Minghao Chen, Iro Laina, and Andrea Vedaldi.
\newblock {DGE:} direct gaussian {3D} editing by consistent multi-view editing.
\newblock In \emph{Computer Vision - {ECCV} 2024 - 18th European Conference, Milan, Italy, September 29-October 4, 2024, Proceedings, Part {LXXIV}}, pages 74--92. Springer, 2024{\natexlab{a}}.

\bibitem[Chen et~al.(2010)Chen, Guan, and Wang]{chen2010approximate}
Yongjian Chen, Tao Guan, and Cheng Wang.
\newblock Approximate nearest neighbor search by residual vector quantization.
\newblock \emph{Sensors}, 10\penalty0 (12):\penalty0 11259--11273, 2010.

\bibitem[Chen et~al.(2024{\natexlab{b}})Chen, Chen, Zhang, Wang, Yang, Wang, Cai, Yang, Liu, and Lin]{DBLP:conf/cvpr/ChenCZWYWCYLL24}
Yiwen Chen, Zilong Chen, Chi Zhang, Feng Wang, Xiaofeng Yang, Yikai Wang, Zhongang Cai, Lei Yang, Huaping Liu, and Guosheng Lin.
\newblock Gaussianeditor: Swift and controllable {3D} editing with gaussian splatting.
\newblock In \emph{{IEEE/CVF} Conference on Computer Vision and Pattern Recognition, {CVPR} 2024, Seattle, WA, USA, June 16-22, 2024}, pages 21476--21485. {IEEE}, 2024{\natexlab{b}}.

\bibitem[Chen et~al.(2024{\natexlab{c}})Chen, Wu, Lin, Harandi, and Cai]{DBLP:conf/eccv/ChenWLHC24}
Yihang Chen, Qianyi Wu, Weiyao Lin, Mehrtash Harandi, and Jianfei Cai.
\newblock {HAC:} hash-grid assisted context for {3D} gaussian splatting compression.
\newblock In \emph{Computer Vision - {ECCV} 2024 - 18th European Conference, Milan, Italy, September 29-October 4, 2024, Proceedings, Part {VII}}, pages 422--438. Springer, 2024{\natexlab{c}}.

\bibitem[Chung et~al.(2024)Chung, Hyun, and Heo]{DBLP:conf/cvpr/ChungHH24}
Jiwoo Chung, Sangeek Hyun, and Jae{-}Pil Heo.
\newblock Style injection in diffusion: {A} training-free approach for adapting large-scale diffusion models for style transfer.
\newblock In \emph{{IEEE/CVF} Conference on Computer Vision and Pattern Recognition, {CVPR} 2024, Seattle, WA, USA, June 16-22, 2024}, pages 8795--8805. {IEEE}, 2024.

\bibitem[Comanici et~al.(2025)Comanici, Bieber, Schaekermann, Pasupat, Sachdeva, Dhillon, Blistein, Ram, Zhang, Rosen, Marris, Petulla, Gaffney, Aharoni, Lintz, Pais, Jacobsson, Szpektor, Jiang, Haridasan, Omran, Saunshi, Bahri, Mishra, Chu, Boyd, Hekman, Parisi, Zhang, Kawintiranon, Bedrax{-}Weiss, Wang, Xu, Purkiss, Mendlovic, Deutel, Nguyen, Langley, Korn, Rossazza, Ram{\'{e}}, Waghmare, Miller, Byrd, Sheshan, Bhardwaj, Janus, Rissa, Horgan, Silver, Wahid, Brin, Raimond, Kloboves, Wang, Gundavarapu, Shumailov, Wang, Pajarskas, Heyward, Nikoltchev, Kula, Zhou, Garrett, Kafle, Arik, Goel, Yang, Park, Kojima, Mahmoudieh, Kavukcuoglu, Chen, Fritz, Bulyenov, Roy, Paparas, Shemtov, Chen, Strudel, Reitter, Roy, Vlasov, Ryu, Leichner, Yang, Mariet, Vnukov, Sohn, Stuart, Liang, Chen, Rawlani, Koh, Co{-}Reyes, Lai, Banzal, Vytiniotis, Mei, and Cai]{DBLP:journals/corr/abs-2507-06261}
Gheorghe Comanici, Eric Bieber, Mike Schaekermann, Ice Pasupat, Noveen Sachdeva, Inderjit~S. Dhillon, Marcel Blistein, Ori Ram, Dan Zhang, Evan Rosen, Luke Marris, Sam Petulla, Colin Gaffney, Asaf Aharoni, Nathan Lintz, Tiago~Cardal Pais, Henrik Jacobsson, Idan Szpektor, Nan{-}Jiang Jiang, Krishna Haridasan, Ahmed Omran, Nikunj Saunshi, Dara Bahri, Gaurav Mishra, Eric Chu, Toby Boyd, Brad Hekman, Aaron Parisi, Chaoyi Zhang, Kornraphop Kawintiranon, Tania Bedrax{-}Weiss, Oliver Wang, Ya Xu, Ollie Purkiss, Uri Mendlovic, Ila{\"{\i}} Deutel, Nam Nguyen, Adam Langley, Flip Korn, Lucia Rossazza, Alexandre Ram{\'{e}}, Sagar Waghmare, Helen Miller, Nathan Byrd, Ashrith Sheshan, Raia Hadsell~Sangnie Bhardwaj, Pawel Janus, Tero Rissa, Dan Horgan, Sharon Silver, Ayzaan Wahid, Sergey Brin, Yves Raimond, Klemen Kloboves, Cindy Wang, Nitesh~Bharadwaj Gundavarapu, Ilia Shumailov, Bo Wang, Mantas Pajarskas, Joe Heyward, Martin Nikoltchev, Maciej Kula, Hao Zhou, Zachary Garrett, Sushant Kafle, Sercan Arik, Ankita Goel,
  Mingyao Yang, Jiho Park, Koji Kojima, Parsa Mahmoudieh, Koray Kavukcuoglu, Grace Chen, Doug Fritz, Anton Bulyenov, Sudeshna Roy, Dimitris Paparas, Hadar Shemtov, Bo{-}Juen Chen, Robin Strudel, David Reitter, Aurko Roy, Andrey Vlasov, Changwan Ryu, Chas Leichner, Haichuan Yang, Zelda Mariet, Denis Vnukov, Tim Sohn, Amy Stuart, Wei Liang, Minmin Chen, Praynaa Rawlani, Christy Koh, JD Co{-}Reyes, Guangda Lai, Praseem Banzal, Dimitrios Vytiniotis, Jieru Mei, and Mu Cai.
\newblock Gemini 2.5: Pushing the frontier with advanced reasoning, multimodality, long context, and next generation agentic capabilities.
\newblock \emph{CoRR}, abs/2507.06261, 2025.

\bibitem[Deng et~al.(2022)Deng, Tang, Dong, Ma, Pan, Wang, and Xu]{DBLP:conf/cvpr/DengTDMPWX22}
Yingying Deng, Fan Tang, Weiming Dong, Chongyang Ma, Xingjia Pan, Lei Wang, and Changsheng Xu.
\newblock Stytr\({}^{\mbox{2}}\): Image style transfer with transformers.
\newblock In \emph{{IEEE/CVF} Conference on Computer Vision and Pattern Recognition, {CVPR} 2022, New Orleans, LA, USA, June 18-24, 2022}, pages 11316--11326. {IEEE}, 2022.

\bibitem[Downs et~al.(2022)Downs, Francis, Koenig, Kinman, Hickman, Reymann, McHugh, and Vanhoucke]{DBLP:conf/icra/DownsFKKHRMV22}
Laura Downs, Anthony Francis, Nate Koenig, Brandon Kinman, Ryan Hickman, Krista Reymann, Thomas~Barlow McHugh, and Vincent Vanhoucke.
\newblock Google scanned objects: {A} high-quality dataset of {3D} scanned household items.
\newblock In \emph{2022 International Conference on Robotics and Automation, {ICRA} 2022, Philadelphia, PA, USA, May 23-27, 2022}, pages 2553--2560. {IEEE}, 2022.

\bibitem[Fan et~al.(2024)Fan, Wang, Wen, Zhu, Xu, and Wang]{DBLP:conf/nips/FanWWZXW24}
Zhiwen Fan, Kevin Wang, Kairun Wen, Zehao Zhu, Dejia Xu, and Zhangyang Wang.
\newblock Lightgaussian: Unbounded {3D} gaussian compression with 15x reduction and 200+ {FPS}.
\newblock In \emph{Advances in Neural Information Processing Systems 38: Annual Conference on Neural Information Processing Systems 2024, NeurIPS 2024, Vancouver, BC, Canada, December 10 - 15, 2024}, 2024.

\bibitem[Fridovich{-}Keil et~al.(2022)Fridovich{-}Keil, Yu, Tancik, Chen, Recht, and Kanazawa]{DBLP:conf/cvpr/Fridovich-KeilY22}
Sara Fridovich{-}Keil, Alex Yu, Matthew Tancik, Qinhong Chen, Benjamin Recht, and Angjoo Kanazawa.
\newblock Plenoxels: Radiance fields without neural networks.
\newblock In \emph{{IEEE/CVF} Conference on Computer Vision and Pattern Recognition, {CVPR} 2022, New Orleans, LA, USA, June 18-24, 2022}, pages 5491--5500. {IEEE}, 2022.

\bibitem[Gao et~al.(2024)Gao, Gu, Lin, Li, Zhu, Cao, Zhang, and Yao]{DBLP:conf/eccv/GaoGLLZCZY24}
Jian Gao, Chun Gu, Youtian Lin, Zhihao Li, Hao Zhu, Xun Cao, Li Zhang, and Yao Yao.
\newblock Relightable {3D} gaussians: Realistic point cloud relighting with {BRDF} decomposition and ray tracing.
\newblock In \emph{Computer Vision - {ECCV} 2024 - 18th European Conference, Milan, Italy, September 29-October 4, 2024, Proceedings, Part {XLV}}, pages 73--89. Springer, 2024.

\bibitem[Ge et~al.(2014)Ge, He, Ke, and Sun]{DBLP:journals/pami/GeHK014}
Tiezheng Ge, Kaiming He, Qifa Ke, and Jian Sun.
\newblock Optimized product quantization.
\newblock \emph{{IEEE} Trans. Pattern Anal. Mach. Intell.}, 36\penalty0 (4):\penalty0 744--755, 2014.

\bibitem[Girish et~al.(2024)Girish, Gupta, and Shrivastava]{DBLP:conf/eccv/GirishGS24}
Sharath Girish, Kamal Gupta, and Abhinav Shrivastava.
\newblock {EAGLES:} efficient accelerated {3D} gaussians with lightweight encodings.
\newblock In \emph{Computer Vision - {ECCV} 2024 - 18th European Conference, Milan, Italy, September 29-October 4, 2024, Proceedings, Part {LXIII}}, pages 54--71. Springer, 2024.

\bibitem[Gray(1984)]{gray1984vector}
Robert Gray.
\newblock Vector quantization.
\newblock \emph{IEEE Assp Magazine}, 1\penalty0 (2):\penalty0 4--29, 1984.

\bibitem[Gu{\'{e}}don and Lepetit(2024)]{DBLP:conf/cvpr/GuedonL24}
Antoine Gu{\'{e}}don and Vincent Lepetit.
\newblock Sugar: Surface-aligned gaussian splatting for efficient {3D} mesh reconstruction and high-quality mesh rendering.
\newblock In \emph{{IEEE/CVF} Conference on Computer Vision and Pattern Recognition, {CVPR} 2024, Seattle, WA, USA, June 16-22, 2024}, pages 5354--5363. {IEEE}, 2024.

\bibitem[Guo et~al.(2024)Guo, Bai, Hu, Guo, Liu, Cai, Huang, and Ma]{DBLP:conf/mm/GuoBHGLC0024}
Yijia Guo, Yuanxi Bai, Liwen Hu, Ziyi Guo, Mianzhi Liu, Yu Cai, Tiejun Huang, and Lei Ma.
\newblock {PRTGS:} precomputed radiance transfer of gaussian splats for real-time high-quality relighting.
\newblock In \emph{Proceedings of the 32nd {ACM} International Conference on Multimedia, {MM} 2024, Melbourne, VIC, Australia, 28 October 2024 - 1 November 2024}, pages 5112--5120. {ACM}, 2024.

\bibitem[Hong et~al.(2023)Hong, Jeon, Lee, Ahn, Kim, Lee, Kim, Uh, and Byun]{DBLP:conf/iccv/HongJLAKLKUB23}
Kibeom Hong, Seogkyu Jeon, Junsoo Lee, Namhyuk Ahn, Kunhee Kim, Pilhyeon Lee, Daesik Kim, Youngjung Uh, and Hyeran Byun.
\newblock Aespa-net: Aesthetic pattern-aware style transfer networks.
\newblock In \emph{{IEEE/CVF} International Conference on Computer Vision, {ICCV} 2023, Paris, France, October 1-6, 2023}, pages 22701--22710. {IEEE}, 2023.

\bibitem[Huber(1992)]{huber1992robust}
Peter~J Huber.
\newblock Robust estimation of a location parameter.
\newblock In \emph{Breakthroughs in statistics: Methodology and distribution}, pages 492--518. Springer, 1992.

\bibitem[Huijben et~al.(2024)Huijben, Douze, Muckley, van Sloun, and Verbeek]{DBLP:conf/icml/HuijbenDMSV24}
Iris A.~M. Huijben, Matthijs Douze, Matthew~J. Muckley, Ruud van Sloun, and Jakob Verbeek.
\newblock Residual quantization with implicit neural codebooks.
\newblock In \emph{Forty-first International Conference on Machine Learning, {ICML} 2024, Vienna, Austria, July 21-27, 2024}. OpenReview.net, 2024.

\bibitem[Jain et~al.(2024)Jain, Kuthiala, Sethi, and Saxena]{jain2024stylesplat}
Sahil Jain, Avik Kuthiala, Prabhdeep~Singh Sethi, and Prakanshul Saxena.
\newblock Stylesplat: {3D} object style transfer with gaussian splatting.
\newblock \emph{arXiv preprint arXiv:2407.09473}, 2024.

\bibitem[J{\'{e}}gou et~al.(2011)J{\'{e}}gou, Douze, and Schmid]{DBLP:journals/pami/JegouDS11}
Herv{\'{e}} J{\'{e}}gou, Matthijs Douze, and Cordelia Schmid.
\newblock Product quantization for nearest neighbor search.
\newblock \emph{{IEEE} Trans. Pattern Anal. Mach. Intell.}, 33\penalty0 (1):\penalty0 117--128, 2011.

\bibitem[Juang and Jr.(1982)]{DBLP:conf/icassp/JuangG82}
Biing{-}Hwang Juang and Augustine H.~Gray Jr.
\newblock Multiple stage vector quantization for speech coding.
\newblock In \emph{{IEEE} International Conference on Acoustics, Speech, and Signal Processing, {ICASSP} '82, Paris, France, May 3-5, 1982}, pages 597--600. {IEEE}, 1982.

\bibitem[Kerbl et~al.(2023)Kerbl, Kopanas, Leimk{\"{u}}hler, and Drettakis]{DBLP:journals/tog/KerblKLD23}
Bernhard Kerbl, Georgios Kopanas, Thomas Leimk{\"{u}}hler, and George Drettakis.
\newblock {3D} gaussian splatting for real-time radiance field rendering.
\newblock \emph{{ACM} Trans. Graph.}, 42\penalty0 (4):\penalty0 139:1--139:14, 2023.

\bibitem[Lee et~al.(2022)Lee, Kim, Kim, Cho, and Han]{DBLP:conf/cvpr/LeeKKCH22}
Doyup Lee, Chiheon Kim, Saehoon Kim, Minsu Cho, and Wook{-}Shin Han.
\newblock Autoregressive image generation using residual quantization.
\newblock In \emph{{IEEE/CVF} Conference on Computer Vision and Pattern Recognition, {CVPR} 2022, New Orleans, LA, USA, June 18-24, 2022}, pages 11513--11522. {IEEE}, 2022.

\bibitem[Lee et~al.(2024)Lee, Rho, Sun, Ko, and Park]{DBLP:conf/cvpr/LeeRSKP24}
Joo~Chan Lee, Daniel Rho, Xiangyu Sun, Jong~Hwan Ko, and Eunbyung Park.
\newblock Compact {3D} gaussian representation for radiance field.
\newblock In \emph{{IEEE/CVF} Conference on Computer Vision and Pattern Recognition, {CVPR} 2024, Seattle, WA, USA, June 16-22, 2024}, pages 21719--21728. {IEEE}, 2024.

\bibitem[Li et~al.(2024)Li, Suzuki, Du, Lee, Atanasov, and Nguyen]{li2024splatsdf}
Runfa~Blark Li, Keito Suzuki, Bang Du, Ki~Myung~Brian Lee, Nikolay Atanasov, and Truong Nguyen.
\newblock Splatsdf: Boosting neural implicit sdf via gaussian splatting fusion.
\newblock \emph{arXiv preprint arXiv:2411.15468}, 2024.

\bibitem[Liu et~al.(2024)Liu, Zhan, Xu, Theobalt, Shao, and Lu]{DBLP:conf/siggrapha/LiuZXT0L24}
Kunhao Liu, Fangneng Zhan, Muyu Xu, Christian Theobalt, Ling Shao, and Shijian Lu.
\newblock Stylegaussian: Instant {3D} style transfer with gaussian splatting.
\newblock In \emph{{SIGGRAPH} Asia 2024 Technical Communications, {SA} 2024, Tokyo, Japan, December 3-6, 2024}, pages 21:1--21:4. {ACM}, 2024.

\bibitem[Liu et~al.(2025)Liu, Li, and Gao]{liu2025image}
Xinshuang Liu, Siqi Li, and Yue Gao.
\newblock Image matting and {3D} reconstruction in one loop.
\newblock \emph{International Journal of Computer Vision}, pages 1--21, 2025.

\bibitem[Loshchilov and Hutter(2017)]{DBLP:conf/iclr/LoshchilovH17}
Ilya Loshchilov and Frank Hutter.
\newblock {SGDR:} stochastic gradient descent with warm restarts.
\newblock In \emph{5th International Conference on Learning Representations, {ICLR} 2017, Toulon, France, April 24-26, 2017, Conference Track Proceedings}. OpenReview.net, 2017.

\bibitem[Loshchilov and Hutter(2019)]{DBLP:conf/iclr/LoshchilovH19}
Ilya Loshchilov and Frank Hutter.
\newblock Decoupled weight decay regularization.
\newblock In \emph{7th International Conference on Learning Representations, {ICLR} 2019, New Orleans, LA, USA, May 6-9, 2019}. OpenReview.net, 2019.

\bibitem[Lu et~al.(2024)Lu, Yu, Xu, Xiangli, Wang, Lin, and Dai]{DBLP:conf/cvpr/0005YXX0L024}
Tao Lu, Mulin Yu, Linning Xu, Yuanbo Xiangli, Limin Wang, Dahua Lin, and Bo Dai.
\newblock Scaffold-gs: Structured {3D} gaussians for view-adaptive rendering.
\newblock In \emph{{IEEE/CVF} Conference on Computer Vision and Pattern Recognition, {CVPR} 2024, Seattle, WA, USA, June 16-22, 2024}, pages 20654--20664. {IEEE}, 2024.

\bibitem[Martinez et~al.(2014)Martinez, Hoos, and Little]{martinez2014stacked}
Julieta Martinez, Holger~H Hoos, and James~J Little.
\newblock Stacked quantizers for compositional vector compression.
\newblock \emph{arXiv preprint arXiv:1411.2173}, 2014.

\bibitem[Metzer et~al.(2023)Metzer, Richardson, Patashnik, Giryes, and Cohen{-}Or]{DBLP:conf/cvpr/MetzerRPGC23}
Gal Metzer, Elad Richardson, Or Patashnik, Raja Giryes, and Daniel Cohen{-}Or.
\newblock Latent-nerf for shape-guided generation of {3D} shapes and textures.
\newblock In \emph{{IEEE/CVF} Conference on Computer Vision and Pattern Recognition, {CVPR} 2023, Vancouver, BC, Canada, June 17-24, 2023}, pages 12663--12673. {IEEE}, 2023.

\bibitem[Mildenhall et~al.(2022)Mildenhall, Srinivasan, Tancik, Barron, Ramamoorthi, and Ng]{DBLP:journals/cacm/MildenhallSTBRN22}
Ben Mildenhall, Pratul~P. Srinivasan, Matthew Tancik, Jonathan~T. Barron, Ravi Ramamoorthi, and Ren Ng.
\newblock Nerf: representing scenes as neural radiance fields for view synthesis.
\newblock \emph{Commun. {ACM}}, 65\penalty0 (1):\penalty0 99--106, 2022.

\bibitem[M{\"{u}}ller et~al.(2022)M{\"{u}}ller, Evans, Schied, and Keller]{DBLP:journals/tog/MullerESK22}
Thomas M{\"{u}}ller, Alex Evans, Christoph Schied, and Alexander Keller.
\newblock Instant neural graphics primitives with a multiresolution hash encoding.
\newblock \emph{{ACM} Trans. Graph.}, 41\penalty0 (4):\penalty0 102:1--102:15, 2022.

\bibitem[Navaneet et~al.(2024)Navaneet, Meibodi, Koohpayegani, and Pirsiavash]{DBLP:conf/eccv/NavaneetMKP24}
K.~L. Navaneet, Kossar~Pourahmadi Meibodi, Soroush~Abbasi Koohpayegani, and Hamed Pirsiavash.
\newblock Compgs: Smaller and faster gaussian splatting with vector quantization.
\newblock In \emph{Computer Vision - {ECCV} 2024 - 18th European Conference, Milan, Italy, September 29-October 4, 2024, Proceedings, Part {XXXII}}, pages 330--349. Springer, 2024.

\bibitem[Niedermayr et~al.(2024)Niedermayr, Stumpfegger, and Westermann]{DBLP:conf/cvpr/NiedermayrSW24}
Simon Niedermayr, Josef Stumpfegger, and R{\"{u}}diger Westermann.
\newblock Compressed {3D} gaussian splatting for accelerated novel view synthesis.
\newblock In \emph{{IEEE/CVF} Conference on Computer Vision and Pattern Recognition, {CVPR} 2024, Seattle, WA, USA, June 16-22, 2024}, pages 10349--10358. {IEEE}, 2024.

\bibitem[Oquab et~al.(2024)Oquab, Darcet, Moutakanni, Vo, Szafraniec, Khalidov, Fernandez, Haziza, Massa, El{-}Nouby, Assran, Ballas, Galuba, Howes, Huang, Li, Misra, Rabbat, Sharma, Synnaeve, Xu, J{\'{e}}gou, Mairal, Labatut, Joulin, and Bojanowski]{DBLP:journals/tmlr/OquabDMVSKFHMEA24}
Maxime Oquab, Timoth{\'{e}}e Darcet, Th{\'{e}}o Moutakanni, Huy~V. Vo, Marc Szafraniec, Vasil Khalidov, Pierre Fernandez, Daniel Haziza, Francisco Massa, Alaaeldin El{-}Nouby, Mido Assran, Nicolas Ballas, Wojciech Galuba, Russell Howes, Po{-}Yao Huang, Shang{-}Wen Li, Ishan Misra, Michael Rabbat, Vasu Sharma, Gabriel Synnaeve, Hu Xu, Herv{\'{e}} J{\'{e}}gou, Julien Mairal, Patrick Labatut, Armand Joulin, and Piotr Bojanowski.
\newblock Dinov2: Learning robust visual features without supervision.
\newblock \emph{Trans. Mach. Learn. Res.}, 2024, 2024.

\bibitem[Qian et~al.(2024)Qian, Wang, Mihajlovic, Geiger, and Tang]{DBLP:conf/cvpr/QianWM0024}
Zhiyin Qian, Shaofei Wang, Marko Mihajlovic, Andreas Geiger, and Siyu Tang.
\newblock {3DGS-Avatar}: Animatable avatars via deformable {3D} gaussian splatting.
\newblock In \emph{{IEEE/CVF} Conference on Computer Vision and Pattern Recognition, {CVPR} 2024, Seattle, WA, USA, June 16-22, 2024}, pages 5020--5030. {IEEE}, 2024.

\bibitem[Reiser et~al.(2021)Reiser, Peng, Liao, and Geiger]{DBLP:conf/iccv/ReiserPL021}
Christian Reiser, Songyou Peng, Yiyi Liao, and Andreas Geiger.
\newblock Kilonerf: Speeding up neural radiance fields with thousands of tiny mlps.
\newblock In \emph{2021 {IEEE/CVF} International Conference on Computer Vision, {ICCV} 2021, Montreal, QC, Canada, October 10-17, 2021}, pages 14315--14325. {IEEE}, 2021.

\bibitem[Ronneberger et~al.(2015)Ronneberger, Fischer, and Brox]{DBLP:conf/miccai/RonnebergerFB15}
Olaf Ronneberger, Philipp Fischer, and Thomas Brox.
\newblock U-net: Convolutional networks for biomedical image segmentation.
\newblock In \emph{Medical Image Computing and Computer-Assisted Intervention - {MICCAI} 2015 - 18th International Conference Munich, Germany, October 5 - 9, 2015, Proceedings, Part {III}}, pages 234--241. Springer, 2015.

\bibitem[Sabin and Gray(2003)]{sabin2003product}
ML Sabin and R Gray.
\newblock Product code vector quantizers for waveform and voice coding.
\newblock \emph{IEEE transactions on acoustics, speech, and signal processing}, 32\penalty0 (3):\penalty0 474--488, 2003.

\bibitem[van~den Oord et~al.(2017)van~den Oord, Vinyals, and Kavukcuoglu]{DBLP:conf/nips/OordVK17}
A{\"{a}}ron van~den Oord, Oriol Vinyals, and Koray Kavukcuoglu.
\newblock Neural discrete representation learning.
\newblock In \emph{Advances in Neural Information Processing Systems 30: Annual Conference on Neural Information Processing Systems 2017, December 4-9, 2017, Long Beach, CA, {USA}}, pages 6306--6315, 2017.

\bibitem[Vaswani et~al.(2017)Vaswani, Shazeer, Parmar, Uszkoreit, Jones, Gomez, Kaiser, and Polosukhin]{DBLP:conf/nips/VaswaniSPUJGKP17}
Ashish Vaswani, Noam Shazeer, Niki Parmar, Jakob Uszkoreit, Llion Jones, Aidan~N. Gomez, Lukasz Kaiser, and Illia Polosukhin.
\newblock Attention is all you need.
\newblock In \emph{Advances in Neural Information Processing Systems 30: Annual Conference on Neural Information Processing Systems 2017, December 4-9, 2017, Long Beach, CA, {USA}}, pages 5998--6008, 2017.

\bibitem[Wang et~al.(2023)Wang, Han, Habermann, Daniilidis, Theobalt, and Liu]{DBLP:conf/iccv/WangHHDTL23}
Yiming Wang, Qin Han, Marc Habermann, Kostas Daniilidis, Christian Theobalt, and Lingjie Liu.
\newblock Neus2: Fast learning of neural implicit surfaces for multi-view reconstruction.
\newblock In \emph{{IEEE/CVF} International Conference on Computer Vision, {ICCV} 2023, Paris, France, October 1-6, 2023}, pages 3272--3283. {IEEE}, 2023.

\bibitem[Wang et~al.(2024{\natexlab{a}})Wang, Li, Guo, Yang, Kot, and Wen]{DBLP:conf/nips/WangLGYKW24}
Yufei Wang, Zhihao Li, Lanqing Guo, Wenhan Yang, Alex~C. Kot, and Bihan Wen.
\newblock Contextgs : Compact {3D} gaussian splatting with anchor level context model.
\newblock In \emph{Advances in Neural Information Processing Systems 38: Annual Conference on Neural Information Processing Systems 2024, NeurIPS 2024, Vancouver, BC, Canada, December 10 - 15, 2024}, 2024{\natexlab{a}}.

\bibitem[Wang et~al.(2024{\natexlab{b}})Wang, Wu, Zhang, and Xu]{DBLP:conf/eccv/WangWZX24}
Yuxin Wang, Qianyi Wu, Guofeng Zhang, and Dan Xu.
\newblock Learning {3D} geometry and feature consistent gaussian splatting for object removal.
\newblock In \emph{Computer Vision - {ECCV} 2024 - 18th European Conference, Milan, Italy, September 29-October 4, 2024, Proceedings, Part {III}}, pages 1--17. Springer, 2024{\natexlab{b}}.

\bibitem[Wang et~al.(2024{\natexlab{c}})Wang, Yi, Wu, Zhao, Chen, and Zhang]{DBLP:conf/eccv/WangYWZCZ24}
Yuxuan Wang, Xuanyu Yi, Zike Wu, Na Zhao, Long Chen, and Hanwang Zhang.
\newblock View-consistent {3D} editing with gaussian splatting.
\newblock In \emph{Computer Vision - {ECCV} 2024 - 18th European Conference, Milan, Italy, September 29-October 4, 2024, Proceedings, Part {XXXV}}, pages 404--420. Springer, 2024{\natexlab{c}}.

\bibitem[Wang et~al.(2004)Wang, Bovik, Sheikh, and Simoncelli]{DBLP:journals/tip/WangBSS04}
Zhou Wang, Alan~C. Bovik, Hamid~R. Sheikh, and Eero~P. Simoncelli.
\newblock Image quality assessment: from error visibility to structural similarity.
\newblock \emph{{IEEE} Trans. Image Process.}, 13\penalty0 (4):\penalty0 600--612, 2004.

\bibitem[Wu et~al.(2024)Wu, Bian, Li, Wang, Reid, Torr, and Prisacariu]{DBLP:conf/eccv/WuBLWRTP24}
Jing Wu, Jia{-}Wang Bian, Xinghui Li, Guangrun Wang, Ian~D. Reid, Philip Torr, and Victor~Adrian Prisacariu.
\newblock Gaussctrl: Multi-view consistent text-driven {3D} gaussian splatting editing.
\newblock In \emph{Computer Vision - {ECCV} 2024 - 18th European Conference, Milan, Italy, September 29-October 4, 2024, Proceedings, Part {XIV}}, pages 55--71. Springer, 2024.

\bibitem[Xu et~al.(2022)Xu, Xu, Philip, Bi, Shu, Sunkavalli, and Neumann]{DBLP:conf/cvpr/XuXPBSSN22}
Qiangeng Xu, Zexiang Xu, Julien Philip, Sai Bi, Zhixin Shu, Kalyan Sunkavalli, and Ulrich Neumann.
\newblock Point-nerf: Point-based neural radiance fields.
\newblock In \emph{{IEEE/CVF} Conference on Computer Vision and Pattern Recognition, {CVPR} 2022, New Orleans, LA, USA, June 18-24, 2022}, pages 5428--5438. {IEEE}, 2022.

\bibitem[Zhang et~al.(2018)Zhang, Isola, Efros, Shechtman, and Wang]{DBLP:conf/cvpr/ZhangIESW18}
Richard Zhang, Phillip Isola, Alexei~A. Efros, Eli Shechtman, and Oliver Wang.
\newblock The unreasonable effectiveness of deep features as a perceptual metric.
\newblock In \emph{2018 {IEEE} Conference on Computer Vision and Pattern Recognition, {CVPR} 2018, Salt Lake City, UT, USA, June 18-22, 2018}, pages 586--595. Computer Vision Foundation / {IEEE} Computer Society, 2018.

\bibitem[Zwicker et~al.(2001)Zwicker, Pfister, van Baar, and Gross]{DBLP:conf/siggraph/ZwickerPBG01}
Matthias Zwicker, Hanspeter Pfister, Jeroen van Baar, and Markus~H. Gross.
\newblock Surface splatting.
\newblock In \emph{Proceedings of the 28th Annual Conference on Computer Graphics and Interactive Techniques, {SIGGRAPH} 2001, Los Angeles, California, USA, August 12-17, 2001}, pages 371--378. {ACM}, 2001.

\end{thebibliography}
    \fi
}

\clearpage
\setcounter{page}{1}
\setcounter{section}{0}
\counterwithin{figure}{section}
\counterwithin{table}{section}
\renewcommand{\thesection}{\Alph{section}}
\maketitlesupplementary

\section{Model Architecture Details}

\begin{figure*}[h]
    \centering
    \includegraphics[width=\linewidth]{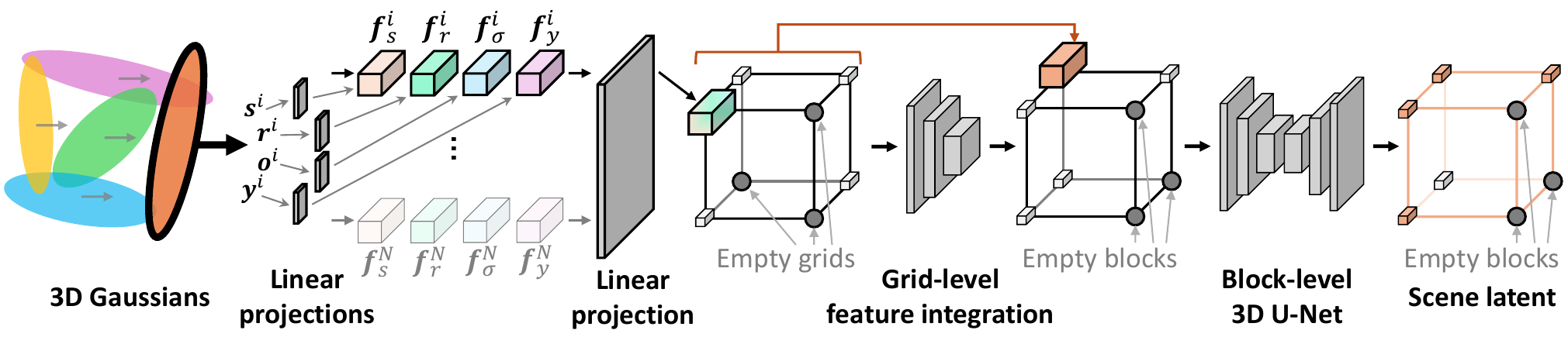}
    \caption{
        \textbf{Scene encoding process in ICGS-Quantizer.}
        For clarity, we visualize a toy configuration with $2{\times}2{\times}2$ grids and blocks.
        For the $i$-th Gaussian, $\boldsymbol{s}^i$, $\boldsymbol{r}^i$, $\sigma^i$, and $\boldsymbol{y}^i$ denote the scaling factor, quaternion, opacity, and spherical harmonics (SH) coefficients, respectively.
        $\boldsymbol{f}_s^i$, $\boldsymbol{f}_r^i$, $\boldsymbol{f}_\sigma^i$, and $\boldsymbol{f}_y^i$ are their corresponding projected feature vectors.
    }
    \label{appendix_fig:encoding}
\end{figure*}

This section provides implementation details to facilitate reproduction of our results, including the encoding process to obtain the scene latent and the visibility estimation used for image conditioning.

\textbf{Encoding process to obtain scene latent.}
The Methodology section in the main paper describes the overall architecture; here we provide a detailed illustration of how we obtain the scene latent (Figure~\ref{appendix_fig:encoding}).
The input to the model is a scene represented by $\{ \mathcal{G}_i \}_{i \in [N]}$. For each Gaussian, we use the attribute scaling factor $\boldsymbol{s}^i$, quaternion $\boldsymbol{r}^i$, opacity $\sigma^i$, and spherical harmonics (SH) coefficients $\boldsymbol{y}^i$ as input. These attributes are individually projected to feature vectors $\boldsymbol{f}_s^i$, $\boldsymbol{f}_r^i$, $\boldsymbol{f}_\sigma^i$, and $\boldsymbol{f}_y^i$ through linear layers. The Gaussian center serves as the grid position but is not used as a feature input. We exclude Gaussians whose opacity logits are below the threshold $\tau_o{=}0$. The four feature vectors are concatenated and linearly projected to a grid feature vector $\boldsymbol{f}_g \in \mathbb{R}^{d_g}$ with $d_g{=}32$, where $d_g{=}32$ denotes the grid-level feature dimensionality. 
To capture local spatial structure, the 3D grid is partitioned into blocks of size $K{\times}K{\times}K$ (with $K{=}4$), and only non‑empty blocks are retained. Within each block, grid features are processed and downsampled by 3D convolutional layers to produce a block feature vector $\boldsymbol{f}_b \in \mathbb{R}^{d_b}$ ($d_b{=}128$). Empty grids are padded with a learnable vector. The set of non‑empty blocks forms a sparse 3D feature map, which is further processed by a block‑level 3D U‑Net with learnable padding to produce a sparse latent scene representation.

\textbf{Visibility estimation.}
For image conditioning, grid‑level image features are obtained by projecting non‑empty grid centers onto the image plane(s) and querying the corresponding feature maps. The retrieved features are then averaged per grid cell, weighted by view-dependent visibility. 
To robustly estimate visibility, we assign to each non-empty grid a virtual sphere centered at the grid position with diameter equal to twice the grid size. For each view, occlusion is computed using this spherical proxy. The visibility of a grid is defined as the visibility of the sphere center for that view. We further propagate visibility at the block level: if any grid within a block is visible from a given view, all grids in that block are marked visible for that view. This strategy reduces sensitivity to local errors in Gaussian shape or opacity.

\section{Training Loss Function Details}

In the Method section of the main paper, we described the training objectives and schedules. This section provides additional details of the loss functions that help reproduce our results. The losses include both Gaussian‑attribute losses, which provide direct guidance early in training, and image losses, which enhance final visual quality.

\textbf{Gaussian-attribute loss.}
To develop a practical model, we train ICGS‑Quantizer on scenes containing real‑world scanned objects. However, such scans often contain noise and high‑frequency textures (\textit{e.g.}, small printed text), which can introduce noisy artifacts in the optimized Gaussians. In addition to automated data cleaning with a vision‑language model (Gemini 2.5 Pro~\cite{DBLP:journals/corr/abs-2507-06261}), we further use robust Gaussian‑attribute objectives that provide straightforward guidance early in geometry learning. Specifically, the Gaussian‑attribute loss $\mathcal{L}_G$ is
\begin{equation}
\mathcal{L}_G = \mathcal{L}_q + \mathcal{L}_s + \mathcal{L}_\sigma,
\end{equation}
where $\mathcal{L}_q$ and $\mathcal{L}_s$ are Huber losses for quaternions and scales, respectively, and $\mathcal{L}_\sigma$ is a binary cross‑entropy (BCE) loss on opacity (from logits). For scales, we apply the Huber objective both in log-space and linear space to balance training stability and accuracy. All terms are weighted equally to minimize the number of hyperparameters and enhance reproducibility. The Huber threshold is set dynamically to the 90th percentile of the current loss values and updated via a moving average.

\textbf{Image loss function.}
For image loss $\mathcal{L}_I$, we adopt the photometric loss from 3DGS~\cite{DBLP:journals/tog/KerblKLD23} and add a mean squared error (MSE) term for training stability:
\begin{equation}
    \mathcal{L}_I = \lambda \mathcal{L}_{\text{D-SSIM}} + \frac{1-\lambda}{2}(\mathcal{L}_1+\mathcal{L}_2),
\end{equation}
with $\lambda{=}0.2$ as in~\cite{DBLP:journals/tog/KerblKLD23}.

\textbf{Final loss.}
We supervise the model at three decoding stages: (i) before image conditioning (to preserve the archived scene), (ii) after \emph{coarse} image conditioning (to adapt to the current scene), and (iii) after \emph{fine} image conditioning (to further refine adaptation). Auxiliary decoders are used accordingly. This design optimizes the quantized codes for both archival fidelity and post-archival adaptability. 
Using the original Gaussian attributes as targets, the final Gaussian-attribute loss is:
\begin{equation}
\mathcal{L}_G^{\text{final}} = \mathcal{L}_G^{\text{scene}} + \mathcal{L}_G^{\text{coarse fusion}}+\mathcal{L}_G^{\text{fine fusion}},
\end{equation}
where $\mathcal{L}_G^{\text{scene}}$ uses the output before image conditioning as the prediction, $\mathcal{L}_G^{\text{coarse fusion}}$ uses the output after \emph{coarse} image conditioning as the prediction, and $\mathcal{L}_G^{\text{fine fusion}}$ uses the output after \emph{fine} image conditioning as the prediction.
Similarly, the final image loss is:
\begin{equation}
\mathcal{L}_I^{\text{final}}=\mathcal{L}_I^{\text{scene}}+\frac{1}{2}\left(\mathcal{L}_I^{\text{coarse fusion}}+\mathcal{L}_I^{\text{fine fusion}}\right),
\end{equation}
where $\mathcal{L}_I^{\text{scene}}$ uses the output before image conditioning as the prediction and the original scene's images as targets, $\mathcal{L}_I^{\text{coarse fusion}}$ uses the output after \emph{coarse} image conditioning as the prediction and the current scene's images as targets, and $\mathcal{L}_I^{\text{fine fusion}}$ uses the output after \emph{fine} image conditioning as the prediction and the current scene's images as targets.
Finally, the total loss is:
\begin{equation}
    \mathcal{L}_{\text{all}} = \mathcal{L}_I^{\text{final}} + w_G \mathcal{L}_G^{\text{final}} + w_{\text{vq}} \mathcal{L}_{\text{vq}},
\end{equation}
where $w_G$ is the weight of the Gaussian-attribute loss, $\mathcal{L}_{\text{vq}}$ is the commitment loss term~\cite{DBLP:conf/nips/OordVK17} for vector quantization and $w_{\text{vq}}$ is its weight. As detailed in the main paper, during epochs 11–90 we linearly reduce $w_G$ from $0.1$ to $0$ while increasing $w_{\text{vq}}$ from $0.01$ to $0.1$.

\section{Implementation Details}

\label{appendix:training}

In addition to the planned release of our code, we provide additional implementation details to support the reproducibility of our research.

\textbf{Codebook learning.} 
We adopt both exponential moving average (EMA) and $k$‑means clustering for codebook learning. EMA updates codebook vectors incrementally at each training step. Every two epochs, $k$‑means clustering is applied to update the codebooks using input vectors computed from the dataset. The codebooks are shared across all training scenes and frozen at test time.

\textbf{Simulating appearance changes.}
To evaluate the capability of our model to adapt scenes to their post-archival changes, we consider each scene to have two states: an archived version at time $t_0$ and a changed version at time $t>t_0$. We consider scene changes in two aspects: (i) the illumination changes and (ii) the appearance changes. To simulate illumination changes, we randomly sample lighting directions and colors. To simulate appearance changes, we randomly rotate the scene texture colors in RGB space around a unit vector. In our scene updating task, the original scene at time $t_0$ serves as the input and the current scene at time $t$ as the target.

\textbf{Number of conditioning images.} 
Our image-conditioned Gaussian splat quantizer can use an arbitrary number of conditioning images. To make the method both robust in scenarios with limited input views and capable of exploiting multi-view information, we train it with a varying number of conditioning images, ranging from 1 to 6 with equal probability. Our training objectives also include supervision for scene recovery without conditioning images, thereby ensuring that the model remains effective even in the absence of input images.

\textbf{Regularization of grid-based 3DGS.} To promote sparse and accurate 3DGS representations, we penalize Gaussian sizes and encourage larger absolute values of Gaussian opacity logits.

\section{Computation of Storage}

This section details how storage is computed for our method and the baselines.

\textbf{Ours.} 
Our method stores each sparse 3D block as $D$ geometry and $D$ texture codes, where $D$ is the number of codebooks used in residual vector quantization. As indices, each code requires $\lceil \log_2 N \rceil$ bits of storage, where $N$ is the codebook size. In addition to the codes, we also store the block index of each block, which requires $\lceil \log_2(B^3) \rceil{=}\lceil 3\log_2 B \rceil$ bits, where $B$ is the block resolution (number of blocks per axis). Thus, each block requires storage of $2D\lceil \log_2N \rceil {+} \lceil 3\log_2B \rceil$ bits. The total storage of one scene is therefore $M\left(2D\lceil \log_2N \rceil {+} \lceil 3\log_2B \rceil\right)$, where $M$ is the number of sparse blocks.

\textbf{Baseline methods.}
For 3DGS~\cite{DBLP:journals/tog/KerblKLD23}, we count all variables as 32‑bit floating‑point numbers, except for the Gaussian centers, which are stored as indices of the grid cells, resulting in less storage.
For C3DGS~\cite{DBLP:conf/cvpr/NiedermayrSW24}, we use the official implementation to compute storage; for each variable, we automatically set the threshold to keep $50\%$ of them at their original values and approximate the remaining $50\%$ by vector quantization.
For CompGS~\cite{DBLP:conf/eccv/NavaneetMKP24}, we use storage computation similar to ours, adding the additional storage of their per‑scene codebooks.

\section{Limitations}
\label{appendix:limitation}

We acknowledge existing limitations of this research. While these aspects lie beyond the primary scope of this work, future improvements could be achieved by exploring additional data compression strategies and optimizing the weighting of loss terms used for training.

\textbf{Comprehensive compression settings.}
To store a scene efficiently, we quantize it into discrete codes represented as integers. Each code requires $\lceil \log_2 N \rceil$ bits when stored as binary data, where $N$ denotes the codebook size. For simplicity and fairness in comparison, we do not apply additional compression techniques to the code data for either our method or the CompGS baseline. However, there remains potential for further improving compression efficiency.
One possible enhancement is Run-Length Encoding (RLE)~\cite{DBLP:conf/eccv/NavaneetMKP24}, which orders the Gaussians according to one of their attributes and stores the frequency of each code value for that attribute, rather than every individual code. 
Another potential improvement is Huffman coding~\cite{DBLP:conf/cvpr/NiedermayrSW24}, which exploits redundancy in the binary representation of the codes by assigning shorter binary sequences to more frequently occurring patterns.

\textbf{Weight of loss functions.} 
In this work, we introduce loss terms to supervise the learning of each Gaussian attribute. The Gaussian attributes for supervision include: (1) a quaternion $\boldsymbol{r} \in \mathbb{R}^4$ representing rotation, (2) a scaling factor $\boldsymbol{s} \in \mathbb{R}^3$, and (3) an opacity value $\sigma \in \mathbb{R}_+$.
In our experiments, we assign each loss term an equal weight of $1$, without performing dataset-specific tuning. This design choice minimizes the number of hyperparameters, thereby enhancing reproducibility. While the overall loss function performs well, there remains potential for further improvement through more effective weighting strategies for the loss terms. One possible direction is to determine optimal weights via hyperparameter searching. Another direction is to automatically balance the loss terms using methods such as Nash Multi-Task Learning.

\end{document}